\documentclass[sigconf]{acmart}

\usepackage{booktabs}
\usepackage{multirow}
\usepackage{amsmath}
\usepackage{url}

\setcopyright{rightsretained}

\acmConference[KDD CUP'19]{ACM SIGKDD Conference on Knowledge Discovery and Data mining}{August 4-8, 2019}{Anchorage, Alaska, USA}
\acmYear{2019}
\copyrightyear{2019}

\begin{document}
\title{Simulating the Effects of Eco-Friendly Transportation Selections \\for Air Pollution Reduction}

\author{Keiichi Ochiai}
\affiliation{\institution{NTT DOCOMO, INC.}}
\authornote{Contact author: ochiaike@nttdocomo.com}

\author{Tsukasa Demizu}
\affiliation{\institution{NTT DOCOMO, INC.}}

\author{Shin Ishiguro}
\affiliation{\institution{NTT DOCOMO, INC.}}

\author{Shohei Maruyama}
\affiliation{\institution{NTT DOCOMO, INC.}}

\author{Akihiro Kawana}
\affiliation{\institution{NTT DOCOMO, INC.}}

\renewcommand{\shortauthors}{K. Ochiai, T. Demizu, S.Ishiguro, S. Maruyama and A. Kawana}

\begin{abstract}
Reducing air pollution, such as CO2 and PM2.5 emissions, is one of the most important issues for many countries worldwide.
Selecting an environmentally friendly transport mode can be an effective approach of individuals to reduce air pollution in daily life.
In this study, we propose a method to simulate the effectiveness of an eco-friendly transport mode selection for reducing air pollution by using map search logs.
We formulate the transport mode selection as a combinatorial optimization problem with the constraints regarding the total amount of CO2 emissions as an example of air pollution and the average travel time.
The optimization results show that the total amount of CO2 emissions can be reduced by 9.23\%, whereas the average travel time can in fact be reduced by 9.96\%.
\end{abstract}

%
%
%


\maketitle

\section{Introduction}
Environmental problem is one of the most important issues in society.
Although many countries worldwide the world are working to reduce air pollution, specifically CO2 emissions,
the United Nations Environment Programme announced that global CO2 emissions have increased in 2017.  
Hence, large-scale actions are necessary to maintain global warming levels below 2$^\circ$C, 
which has been established in the Paris Agreement\footnote{https://www.unenvironment.org/resources/emissions-gap-report-2018}.

Not only global or country-level actions but also individual-level actions are important to address the environmental problem.
One of the approaches for reducing CO2 emissions in daily life is the selection of transportation by each individual.
Individuals can contribute to the reduction of CO2 emissions by opting to environmentally friendly transport mode considering 
that it does not adversely affect their social lives. 
To this end, there are two requirements:
(1) Collecting data on the transport mode of each individual and simulating the effect of transport mode selection for CO2 emission, and 
(2) Allowing a user to choose an eco-friendly mode of transportation by recommending transport mode with low CO2 emission.

With the spread of smartphones, recording mobility behavior on a personal level has become possible by regularly using Global Positioning System (GPS).
Although we can calculate the effects of transport mode selection for CO2 emissions using these data, 
it is difficult to collect large-scale data to track people's locations through GPS constantly for national and government estimates because there is less incentive for individuals.
In addition, services and researches are available to estimate the transport mode from information measured via GPS 
such as Google Maps timeline\footnote{https://www.google.com/maps/timeline}, but accuracy is limited \cite{Macarulla:2018,Zheng:2010}

Assuming that a user has actually moved according to the search results, 
we can identify the coordinates of the original and destination (OD) points, and then transport mode between OD points by using map search logs.
Fortunately, the map search logs provided by Baidu provide a unique opportunity for us to tackle the problem.

In this study, we propose a method to simulate the effects of an eco-friendly transportation selection to reduce CO2 emissions.
Specifically, we formulate the transportation selection as a combinatorial optimization problem with the total amount of travel time and CO2 emission.
We simulate the CO2 emission reduction based the combinatorial optimization.
The simulation results reveal that the total amount of CO2 emissions can be reduced by 9.23\%, and the average travel time can be particularly decreased by 9.96\%.

\section{Approach}\label{sec:approach}
In this section, we first explain the estimation of the actual transport mode because 
the map search logs provided by Baidu do not specify which transport mode corresponds to which actual transport means.
Next, we provide the formulation of transport mode selection with the constraints of the total amount of travel time and CO2 emission.
Then, we elucidate the calculation of CO2 emissions and travel time, and then the equations of constraints.
In our approach, we use the data of query, display and click records described in the KDD CUP page\footnote{https://dianshi.baidu.com/competition/29/question}.
Owing to the space limitation, we omit the explanation of these data. Thus, please refer to the KDD CUP web page for more detail.
In addition, we assume that a user actually traveled by the transport mode in click record although we cannot determine his/her actual transport mode.

\subsection{Estimation of the Actual Transportation Means}
The map search logs provided by Baidu do not specify which transport mode corresponds to which actual transportation means.
Thus, we must estimate the correspondence.

First, we performed a route search in Beijing with the Baidu Map app\footnote{https://play.google.com/store/apps/details?id=com.baidu.BaiduMap} and confirmed what transportation means was displayed.
The transportation means that we could confirm are the following: walking, bicycle, driving, bus, subway, taxi, share cycle and bus, share cycle and subway, 
taxi and bus, taxi and subway, bus and subway, and transportation network app (e.g. DiDi)
We estimated that the 11 transport modes according to the map search logs correspond to 11 of the 12 transportation means described above.

Next, we estimated the correspondence between the transport mode and actual transportation means from the assumed route distance, 
the estimated time of arrival (ETA), and price of each transport mode.
Transport modes 3, 5 and 6 have no information on the estimated price, and the average value of the estimated route distance divided by ETA (i.e. estimated speed) increase in the order of transport modes 5, 6, and 3.
Thus, we estimated that transport mode 3 corresponds to driving, 5 to walking, and 6 to bicycle.
Moreover, we estimate that transport modes 2 and 4 correspond to the subway and taxi respectively,
because the relationship between the estimated distance and the estimated price of transport modes 2 and 4 is similar to the subway and taxi fares in Beijing\footnote{https://www.travelchinaguide.com/cityguides/beijing/transportation/}, respectively.
The estimated price of transport mode 7 is often between bus and subway fares in Beijing, then we estimate that transport mode 7 corresponds to bus and subway.
The estimated prices of transport modes 1 and 11 are similar to the bus fare in Beijing; hence, we consider that they correspond to either bus or \textit{share cycle and bus.}
Considering that transport mode 1 is displayed more often than transport mode 11 in the map search logs, we estimate that transport mode 1 is bus, which is simpler and more easily searchable, and that transport mode 11 is share cycle and bus.

Summarizing these estimation results, we suppose that transport mode 1 corresponds to bus, 2 to subway, 3 to driving, 4 to taxi, 5 to walking, 6 to bicycle, 7 to bus and subway, and 11 to bus and share cycle, respectively.
We do not use the data of transport mode 8, 9 and 10 data in the following discussion because we could not reliably estimate what the actual transportation means they correspond to from the map search logs.

\subsection{Problem Formulation}
The goal of this study is to find the transport mode selection with minimizing the total amount of travel time and CO2 emission.
Given display and click records regarding a specific period, we can formulate the problem by minimizing the following objective function as 0-1 integer programming problem.
\begin{eqnarray}\label{problem}
\min \sum_{i=1}^m \sum_{j=1}^n (P_{i,j}X_{i,j} + Q_{i,j}X_{i,j})
\end{eqnarray}
\begin{eqnarray}
\textrm{subject to} \sum_{j=1}^n X_{i,j} = 1 \\
X_{i,j} \in \{0 ,1\} \\
Q_{i,j}X_{i,j} \leq Q'_{i,j} 
\end{eqnarray}
Here, $m$ is the number of session IDs in click records, $n$ is the number of transport mode,
$P_{i,j}$ is the CO2 emissions for the transport mode $j$ of user $i$, $Q_{i,j}$ is the travel time for transport mode $j$ of user $i$, 
$Q'_{i,j}$ is the travel time of clicked transport mode $j$ of user $i$,
and $X_{i,j}$ indicates the selected transport mode defined as follows.
\[
  X_{i,j} = \begin{cases}
    1 & ( \rm{user\ } \it{i} \ \rm{selected \ transport \ mode \ } \it{j}) \\
    0 & (otherwise)
  \end{cases}
\]

\subsection{Calculation of the CO2 Emissions and Travel Time}
We calculate the CO2 emissions of each transport mode of each user based on CO2 emission per unit distance 
according to the annual report of the Foundation for Promoting Personal Mobility and Ecological Transportation \cite{Tej:2018}
which is supervised by the Ministry of Land, Infrastructure, Transport and Tourism, Japan.
Because each display record contains the distance from the origin to the destination, 
we can calculate the assumed CO2 emission of each transport mode of each user $P_{i,j}X_{i,j}$ by multiplying the distance by the CO2 emission shown in Table \ref{co2_per_km}.
In the case of the combined transport mode such as \textit{share cycle and bus}, we approximated by averaging the corresponding values of the two transport modes.
For example, in the case of \textit{share cycle and bus}, $(66 + 0)/2 = 33 (g/person/km)$ is used for optimization calculation.

For travel time $Q_{i,j}X_{i,j}$, we simply look up the estimated time of arrival (ETA) from each display record.
In addition, we can obtain $Q'_{i,j}$, which indicates the travel time of the user selected transport mode, by combining the display record and click record based on the session ID.

\begin{table}[tb]
\caption{The amount of CO2 emission for transporting a person for 1km \cite{Tej:2018}}
\label{co2_per_km}
\vspace{-3mm}
\begin{tabular}{|l|r|}
\hline
Transport Mode & \multicolumn{1}{l|}{CO2 emission (g/person/km)} \\ \hline
Passenger cars & 145                                             \\ \hline
Bus            & 66                                              \\ \hline
Railway        & 20                                              \\ \hline
Cycling        & 0                                               \\ \hline
Walking        & 0                                               \\ \hline
\end{tabular}
\end{table}

\subsection{Equations of Constraints}
In this section, we explain the ideas behind the equations of constraints.
Equation (2) indicates that a user can select only one transport mode,
Equation (3) denotes that each $X_{i,j}$ is a binary variable,
and Equation (4) imposes the constraint that the optimized travel time is less than or equal to the travel time that appeared in click record (originally selected by a user)
because if this constraint is not imposed then walking or cycling will be assigned to all routes.

\subsection{Solving Combinatorial Optimization Problem}
To solve 0-1 integer programming problem, we can use a commercial or open-source general purpose solver.
In this study, we used the PuLP library \cite{Mitchell:2011}, which is an open source package developed by the University of Auckland.
Our implementation is available at the webpage\footnote{https://qiita.com/keiichi\_ochiai/items/09557c23291d49e7c510}

\section{Evaluation and Discussion}
\subsection{Evaluation Metric and Data}
To simulate the effect of an environmentally-friendly transport mode selection for reducing CO2 emission,
we compared the total amount of CO2 emissions and travel times between a transport mode originally selected by a user (i.e. click record, referred to as baseline)
and a transport mode optimized using our proposed method (referred to as optimized mode) as a quantitative evaluation.
We calculated the total amount of CO2 emissions and travel times of the baseline and optimized mode
based on the data of query, display and click records from October 1, 2018 to November 30, 2018 from Beijing, China.
These data were filtered by the transport modes 1, 2, 3, 4, 5, 6, 7, and 11 which were estimated in Section 2.1.
The total number of session IDs (records) was 387,733.

\subsection{Evaluation Results}\label{evaluation_results}
Table \ref{compare_co2} presents the result of the total amount of CO2 emission, which was reduced by 9.23\% from the baseline to the optimized mode.
Table \ref{compare_time} summarizes the results of the travel time of the baseline and optimized mode.
Surprisingly, the average travel time of the optimized mode was reduced by 9.96\% when compared with that of the baseline.

\begin{table}[tb]
\caption{Comparison of CO2 emissions}
\label{compare_co2}
\vspace{-3mm}
\begin{tabular}{|l|r|r|r|r|}
\hline
Method         & \multicolumn{1}{l|}{Total (t)} & \multicolumn{1}{l|}{Avg (g)} & \multicolumn{1}{l|}{Median (g)} & \multicolumn{1}{l|}{S.D.} \\ \hline
Baseline       & 323.58                         & 834.56                       & 399.26                          & 1346.89                   \\ \hline
Optimized Mode & 293.75                         & 757.61                       & 309.58                          & 1359.55                   \\ \hline
\end{tabular}
\end{table}

\begin{table}[tb]
\caption{Comparison of travel time}
\label{compare_time}
\vspace{-3mm}
\begin{tabular}{|l|r|r|r|}
\hline
Method         & \multicolumn{1}{l|}{Avg. (s)} & \multicolumn{1}{l|}{Median (s)} & \multicolumn{1}{l|}{S.D.} \\ \hline
Baseline       & 3150.33                       & 2734.00                         & 2245.47                   \\ \hline
Optimized Mode & 2836.69                       & 2387.00                         & 2327.67                   \\ \hline
\end{tabular}
\end{table}

The number of counts of transport mode change is shown in Table \ref{compare_mode}. 
In the column of Mode, the left side represents the baseline transport mode and the right side indicates the optimized transport mode.
As observed in the Table \ref{compare_mode}, the transport mode often changes from bus to cycling and from subway to cycling.
Meanwhile, the number of cases from driving to cycling is less than those of the transport mode, which does not change (i.e., from driving to driving).

\begin{figure}[tb]
  \centering
  \includegraphics[width=0.4\textwidth]{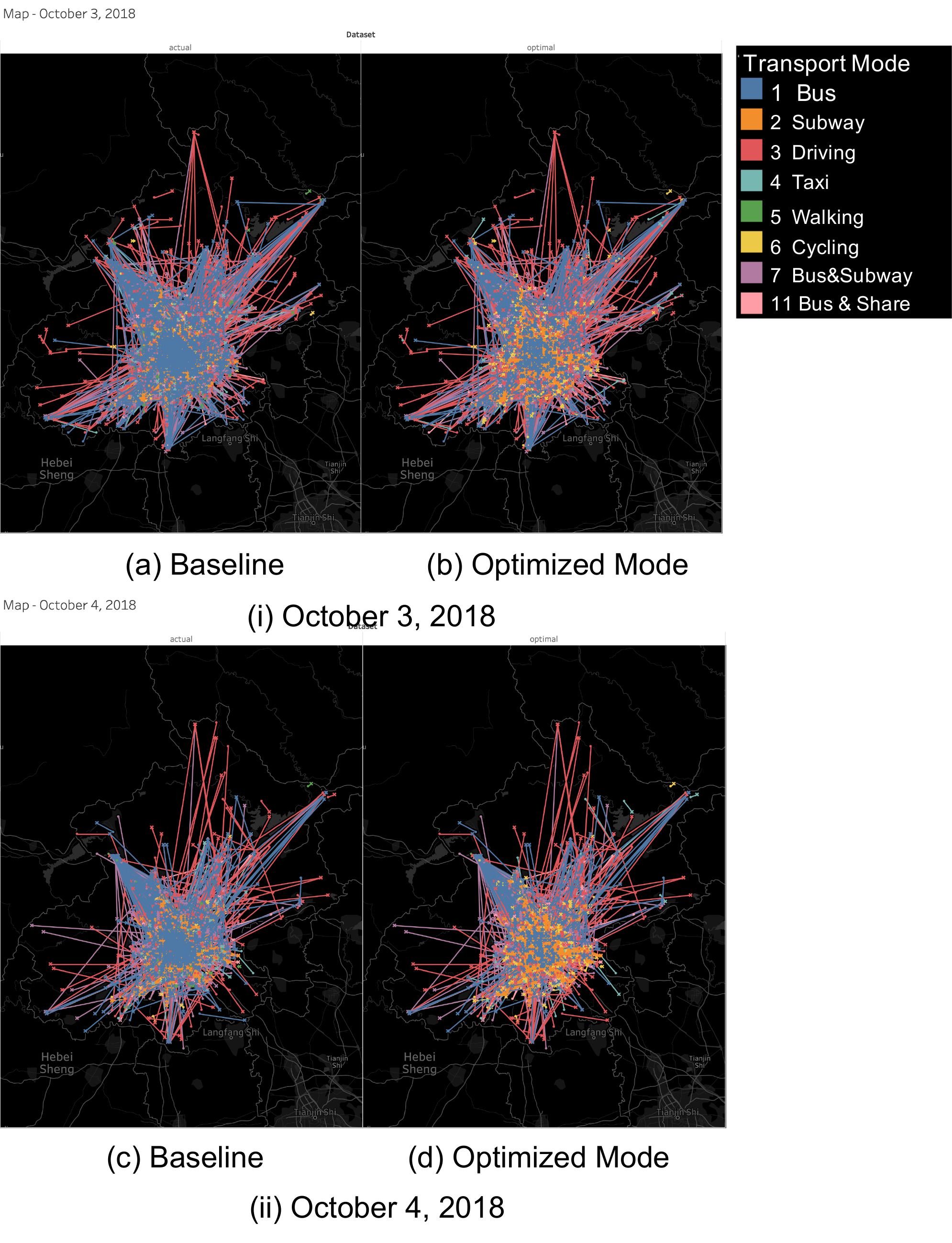}
  \caption{Visualization of the transport mode change on the map}
  \label{fig:visualize}
\end{figure}

Table \ref{compare_example} shows the examples of the change in the transport mode for the baseline and optimized mode. 
In addition, we visualize the change in the transport mode for the baseline and optimized mode for October 3 and 4, 2018, in Figure \ref{fig:visualize}. 
From these results, it may be considered that when the travel distance is relatively short, cycling is the best alternative for both the CO2 emissions and travel time. 
Meanwhile, there may be no other option to arrive at the location by driving (car) in the case of long-distance travel. These results seem reasonable.

\begin{table*}[tb]
\caption{Comparison of the number of each transport mode (selected representative cases)}
\label{compare_mode}
\vspace{-3mm}
\scalebox{0.75}{
\begin{tabular}{|l|r|l|l|r|l|l|r|l|l|r|}
\cline{1-2} \cline{4-5} \cline{7-8} \cline{10-11}
Mode                    & \multicolumn{1}{l|}{Count} &  & Mode                       & \multicolumn{1}{l|}{Count} &  & Mode                        & \multicolumn{1}{l|}{Count} &  & Mode                        & \multicolumn{1}{l|}{Count} \\ \cline{1-2} \cline{4-5} \cline{7-8} \cline{10-11} 
(Bus, Bus)              & 20834                      &  & (Subway, Bus)              & 167                        &  & (Driving, Bus)              & 6                          &  & (Cycling, Bus)              & 3                          \\ \cline{1-2} \cline{4-5} \cline{7-8} \cline{10-11} 
(Bus, Subway)           & 2513                       &  & (Subway, Subway)           & 107015                     &  & (Driving, Subway)           & 325                        &  & (Cycling, Subway)           & 63                         \\ \cline{1-2} \cline{4-5} \cline{7-8} \cline{10-11} 
(Bus, Driving)          & 2786                       &  & (Subway, Driving)          & 3964                       &  & (Driving, Driving)          & 22017                      &  & (Cycling, Driving)          & 58                         \\ \cline{1-2} \cline{4-5} \cline{7-8} \cline{10-11} 
(Bus, Taxi)             & 325                        &  & (Subway, Taxi)             & 184                        &  & (Driving, Taxi)             & 1051                       &  & (Cycling, Taxi)             & 1                          \\ \cline{1-2} \cline{4-5} \cline{7-8} \cline{10-11} 
(Bus, Walking)          & 167                        &  & (Subway, Walking)          & 62                         &  & (Driving, Walking)          & 31                         &  & (Cycling, Walking)          & 21                         \\ \cline{1-2} \cline{4-5} \cline{7-8} \cline{10-11} 
(Bus, Cycling)          & 37532                      &  & (Subway, Cycling)          & 24858                      &  & (Driving, Cycling)          & 1157                       &  & (Cycling, Cycling)          & 11715                      \\ \cline{1-2} \cline{4-5} \cline{7-8} \cline{10-11} 
(Bus, Bus\&Subway)      & 4880                       &  & (Subway, Bus\&Subway)      & 115                        &  & (Driving, Bus\&Subway)      & 36                         &  & (Cycling, Bus\&Subway)      & 2                          \\ \cline{1-2} \cline{4-5} \cline{7-8} \cline{10-11} 
(Bus, Bus\&Share Cycle) & 1332                       &  & (Subway, Bus\&Share Cycle) & 126                        &  & (Driving, Bus\&Share Cycle) & 3                          &  & (Cycling, Bus\&Share Cycle) & 0                          \\ \cline{1-2} \cline{4-5} \cline{7-8} \cline{10-11} 
\end{tabular}
}
\end{table*}

\begin{table}[tb]
\caption{Comparison of the transport mode (selected representative cases)}
\label{compare_example}
\vspace{-3mm}
\begin{center}
\scalebox{0.75}{
\begin{tabular}{|l|l|r|r|l|r|r|}
\hline
        & \multicolumn{3}{c|}{baseline}                                                                                                                                                                                          & \multicolumn{3}{c|}{optimized mode}                                                                                                                                                                                    \\ \hline
sid     & \begin{tabular}[c]{@{}l@{}}transport \\ mode\end{tabular} & \multicolumn{1}{l|}{\begin{tabular}[c]{@{}l@{}}CO2 \\ emission\end{tabular}} & \multicolumn{1}{l|}{\begin{tabular}[c]{@{}l@{}}travel \\ time\end{tabular}} & \begin{tabular}[c]{@{}l@{}}transport \\ mode\end{tabular} & \multicolumn{1}{l|}{\begin{tabular}[c]{@{}l@{}}CO2 \\ emission\end{tabular}} & \multicolumn{1}{l|}{\begin{tabular}[c]{@{}l@{}}travel \\ time\end{tabular}} \\ \hline
2848914 & Bus                                                       & 3508.296                                                                     & 6456                                                                        & Bus                                                       & 3508.296                                                                     & 6456                                                                        \\ \hline
2318006 & Bus                                                       & 228.162                                                                      & 1710                                                                        & Cycling                                                   & 0                                                                            & 846                                                                         \\ \hline
2437983 & Bus                                                       & 494.406                                                                      & 2104                                                                        & Cycling                                                   & 0                                                                            & 1958                                                                        \\ \hline
2869206 & Bus                                                       & 230.604                                                                      & 1960                                                                        & Cycling                                                   & 0                                                                            & 1090                                                                        \\ \hline
1612703 & Bus                                                       & 2425.236                                                                     & 7679                                                                        & Bus\&Subway                                               & 1879.315                                                                     & 6975                                                                        \\ \hline
\end{tabular}
}
\end{center}
\end{table}


\subsection{Discussion}
\subsubsection{Simulating the Effect of Whole China}
As described in Section \ref{evaluation_results}, the total amount of CO2 emissions can be reduced by 9.23\%.
We roughly estimate the effect on the entire country of China based on this result.
According to the report of the International Energy Agency\cite{CO2_fuel}, 
The second-largest sector of global CO2 emissions from fuel combustion is transport (24\%),
and the total amounts of CO2 emissions regarding transport for worldwide and China are 7,737.8 and 843.9 million tons, respectively.
If the total amount of CO2 emissions in China can be reduced by 9.23\%, then 77.9 million tons of CO2 emissions might be reduced.

\subsubsection{Impact on Public Health Perspective}
The increase in the number of people riding bicycles instead of buses and subways is thought to affect not only CO2 emissions but also life expectancy.
Kelly et al. \cite{Kelly:2014} reported that cycling has a positive effect on reducing the risk of all-cause mortality.
WHO recommends that people should engage in physical activity of 11.25 METh/week. If people follow this recommendation, 
then the risk reduction for all-cause mortality becomes 10\% for cycling compared with no cycling activity.
From our simulation, the average duration of cycling was 23.04 min/day. 
Riding a bicycle daily for 23.04 min is 1.53 METh/week which corresponds to 13.63\% of the WHO recommendation.
However, we need additional information to rigorously calculate the risk reduction for our case
because we only estimate that the risk reduction for all-cause mortality may be $10\% \times 13.63\% = 1.36\%$ for this case.

\section{Limitation and Implication}
We require additional data to complete this study thoroughly.
Our study has several limitations as follows.\newline
(1) {\bf Actual transport mode is not identified}.  As mentioned in the beginning of Section \ref{sec:approach}, 
we cannot identify the user move by the transportation recorded in the click records.
If we can obtain continuous GPS trajectories, then we can simulate the effects more accurately.\newline
\noindent(2) {\bf Exact transport mode is required}. Because we estimated each transport mode based on publicly available data, 
the estimated transport mode may be incorrect, leading to inaccurate simulation results.
In addition, we need to calculate the effect based on each transport mode because several transport modes are mixed.\newline
\noindent(3) {\bf We ignored the constraint for the limited number of simultaneous users for each mode}. 
Buses and subways are often replaced by bicycles; however, in reality, due to the number of shared bicycles and the parking space limitations, 
it is not possible for everyone to use a bicycle. Conversely, we may be able to simulate the number of parking spaces for bicycles from our results.\newline
\noindent(4) {\bf We need an UX/UI design to select the eco-friendly transportation}. In this study, we only solve part of the requirement (only requirement 1)
mentioned in the introduction.
To make a user choose an eco-friendly transportation, there are several possible approach.
One possible solution is the improvement of ranking which is related to UI.
Another possible solution is to incorporate the eco-friendly actions into social credit systems such as Zhima Credit\footnote{https://www.xin.xin/}
to motivate them to select an eco-friendly transportation.

\section{Related Work}
Our study is related to map search log analysis. Because companies that have large-scale map search logs are limited, only a few related work for analyzing map search logs exist.
Xiao et al. \cite{Xiao:2010} analyzed the characteristics of map search queries regarding Microsoft Live Maps. 
They found several unique characteristics; for example, users input longer queries and fix queries more frequently.
Kumar et al. \cite{Kumar:2015} explored  the search logs of Google Maps, and found a relationship between the choice of shop and distance for destination and the rank.
To the best of our knowledge, there is no work to simulate the environmental effects and discusses these effects on public health has been conducted yet.

\section{Conclusion}
In this work, we simulated the effects of an eco-friendly transport mode selection for CO2 emissions reduction.
We formulated the transport mode selection as a combinatorial optimization problem with total amount of travel time and CO2 emissions.
In future work, it is expected to be applied to other types of environmental metrics such as PM2.5, and energy consumption etc.

\begin{acks}
We would like to thank Baidu for providing the map search logs.
\end{acks}

\bibliographystyle{ACM-Reference-Format}
\bibliography{reference}

\end{document}